\documentclass[12pt]{iopart}

\expandafter\let\csname equation*\endcsname\relax
\expandafter\let\csname endequation*\endcsname\relax
\usepackage{amssymb}
\usepackage{amsmath}
\usepackage{graphicx}
\usepackage{booktabs}
\usepackage{multirow}
\usepackage{tabularx}
\usepackage{bm}
\usepackage{color}
\usepackage{cite}
\usepackage{ulem}
\usepackage{url}
\usepackage{subcaption}

\newcommand{\revision}[1]{#1}

\begin{document}

\title[Human-AI Collaborative Multi-modal Multi-rater Learning]{Human-AI Collaborative Multi-modal Multi-rater Learning for Endometriosis Diagnosis}

\author{Hu Wang, David Butler, Yuan Zhang, Jodie Avery, Steven Knox, Congbo Ma, Louise Hull, Gustavo Carneiro}

\address{The University of Adelaide, Adelaide, Australia \\
Benson Radiology, Adelaide, Australia \\
Macquarie University, Sydney, Australia \\
University of Surrey, Guildford, United Kingdom}
\begin{indented}
\item[]
\end{indented}

\begin{abstract}
Endometriosis, affecting about 10\% of individuals assigned female at birth, is challenging to diagnose and manage. Diagnosis typically involves the identification of various signs of the disease using either laparoscopic surgery or the analysis of T1/T2 MRI images, with the latter being quicker and cheaper but less accurate. A key diagnostic sign of endometriosis is the obliteration of the Pouch of Douglas (POD). However, even experienced clinicians struggle with accurately classifying POD obliteration from MRI images, which complicates the training of reliable AI models. In this paper, we introduce the \underline{H}uman-\underline{AI} \underline{Co}llaborative \underline{M}ulti-modal \underline{M}ulti-rater Learning (HAICOMM) methodology to address the challenge above. HAICOMM is the first method that explores three important aspects of this problem: 1) multi-rater learning to extract a cleaner label from the multiple ``noisy'' labels available per training sample; 2) multi-modal learning to leverage the presence of T1/T2 MRI images for training and testing; and 3) human-AI collaboration to build a system that leverages the predictions from clinicians and the AI model to provide more accurate classification than standalone clinicians and AI models. Presenting results on the multi-rater T1/T2 MRI endometriosis dataset that we collected to validate our methodology, the proposed HAICOMM model outperforms an ensemble of clinicians, noisy-label learning models, and multi-rater learning methods.
\end{abstract}

%
%
%
%
%

\section{Introduction}

Endometriosis is characterized by the abnormal growth of endometrial-like tissue outside the uterus, often leading to distressing symptoms such as chronic pain, prolonged menstrual bleeding, and infertility~\cite{lagana2020evaluation,moss2021delayed}. Despite its prevalence in around 10\% of individuals assigned female at birth~\cite{australian2019endometriosis}, diagnosing endometriosis has been a hard condition to diagnose.
Conventional diagnostic methods primarily rely on invasive laparoscopy, a surgical procedure that involves the insertion of a slender camera through a small incision in the abdomen to visually inspect the pelvic region~\cite{becker2022eshre}. 
This diagnostic method, while effective, presents substantial drawbacks. 
Chief among them is the significant delay (averaging  6.4 years~\cite{australian2019endometriosis}) that patients endure before receiving a formal diagnosis. This long waiting period lowers the quality of life for those afflicted by the condition~\cite{horne2022pathophysiology}. 
Furthermore, the extensive reliance on laparoscopy escalates healthcare costs, imposing a considerable burden on both healthcare systems and patients~\cite{soliman2016direct}. These challenges underscore the pressing need for innovative imaging-based diagnostic solutions that can mitigate these issues while enhancing patient care.

\begin{figure}[h!]
\centering
\includegraphics[width=0.6\textwidth]{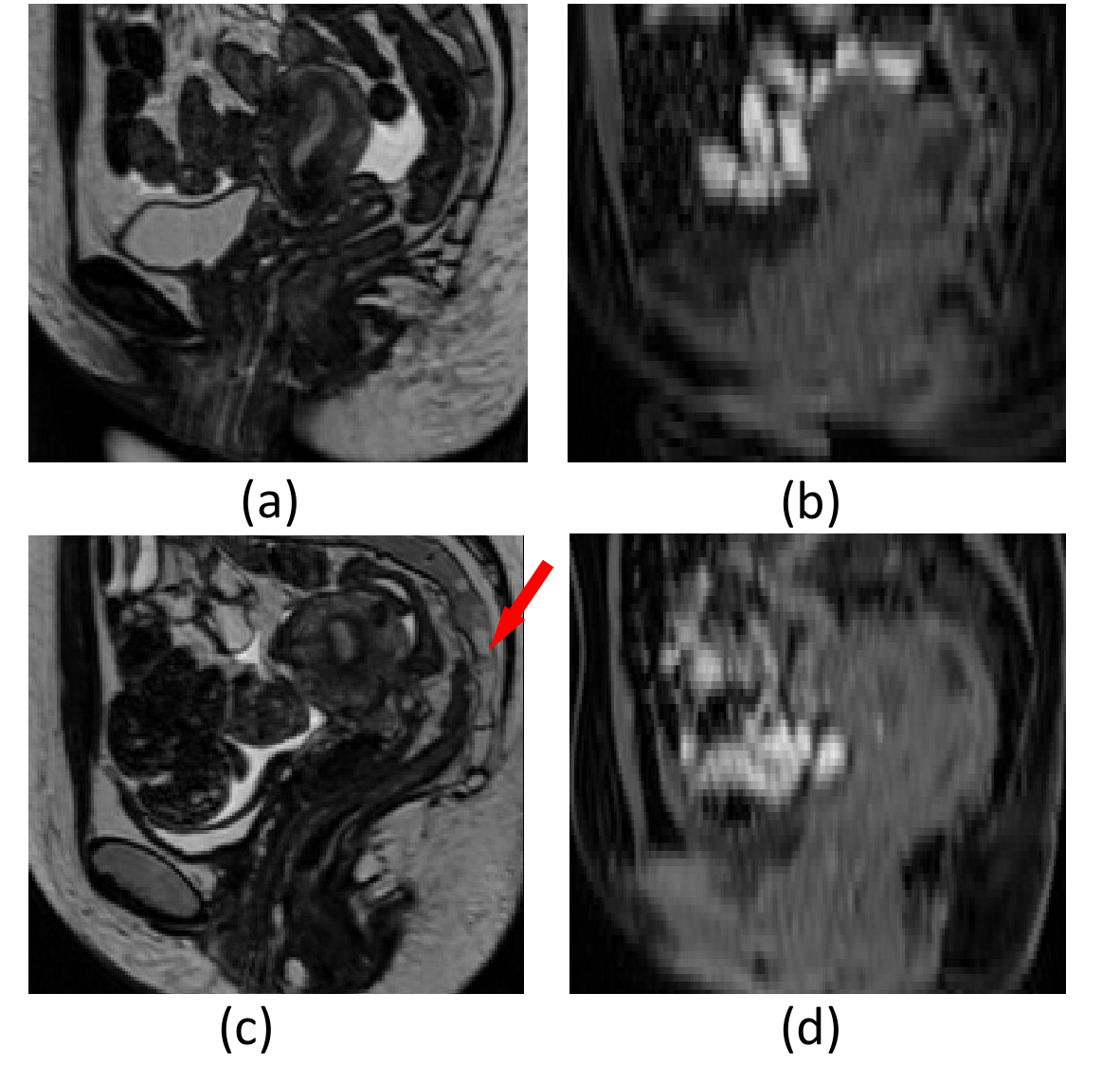}
\caption{\revision{Paired sagittal MR images, with T1-weighted (b, d) and T2-weighted (a, c) scans. The (a, b) pair shows a normal POD, while the (c, d) pair reveals POD obliteration, highlighted by a red arrow indicating significant adhesion and tissue distortion, demonstrating the loss of the soft tissue plane separating the uterine fundus from the bowel.}
}
\label{fig:qualitative_pod}
\end{figure}

The T1 and T2 modalities of Magnetic Resonance Imaging (MRI) are among the most recommended medical imaging methods for diagnosing endometriosis given their effectiveness to visualize many signs of the condition.  One of the most important signs associated with the condition is the 
Pouch of Douglas (POD) Obliteration~\cite{kinkel2006diagnosis,Butler2023TheEO}. 
Developing an AI model capable of classifying POD obliteration has the potential to facilitate the widespread adoption of imaging-based diagnosis and enhance diagnosis accuracy and consistency. 
However, training such a model relies on acquiring precise POD Obliteration annotations from T1/T2 MRIs, which is a challenging task because even experienced clinicians may lack certainty regarding the presence of the sign. 
In fact, the uncertainty in the manual POD obliteration classification from T1/T2 MRI is remarkably low, with only 61.4\% to 71.9\% accuracy~\cite{kataoka2005posterior, indrielle2020diagnostic}. 
\revision{To demonstrate the difficulty in the detection of POD obliteration, we present  Figure~\ref{fig:qualitative_pod}, showing T1-weighted (b) and (d) paired with T2-weighted (a) and (c) MR images in the sagittal plane. The (a) and (b) pair shows a normal POD case, while the (c) and (d) pair shows POD obliteration, with the red arrow pointing to significant adhesion and distortion, indicating the loss of the soft tissue plane separating the uterine fundus from the bowel in the POD.}
\revision{Nevertheless, there have been some attempts to train machine learning models for POD obliteration detection~\cite{Zhang_2023, Butler2023TheEO}. Despite their reported accuracy, these methods have not been validated against ground truth labels from surgical reports, making it challenging to assess their clinical value in the detection of POD obliteration.} 
Therefore, a major research question in this problem is if it is possible to design innovative training and testing methodologies that can lead to highly accurate and clinically useful POD obliteration classification results.

There are many important aspects of this problem that can be leveraged in order to formulate an innovative solution to produce an accurate POD obliteration classifier. 
First, the uncertain manual classification by clinicians can lead to training sets that  contain multiple ``noisy'' labels per training sample (with each label being produced by a different clinician), which can be explored by multi-rater learning mechanisms~\cite{goh2022CrowdLab}.
Second, given that clinicians and AI models may not be highly accurate, the combination of their predictions may lead to more accurate predictions -- such idea is studied by human-AI collaborative classification~\cite{complement_wilder}.
Third, similarly to previous approaches~\cite{Zhang_2023,Butler2023TheEO}, it is important to explore the complementarity of the multiple MRI modalities.

In this paper, we explore the three points listed above to propose the innovative 
 \underline{H}uman-\underline{AI} \underline{Co}llaborative \underline{M}ulti-modal \underline{M}ulti-rater Learning (HAICOMM) methodology.
HAICOMM is the first method in the field that simultaneously explores multi-rater learning to provide a clean training label from the multiple ``noisy'' labels produced by clinicians, multi-modal learning to leverage the presence of T1/T2 MRI images, and human-AI collaboration to build a system 
that synergises predictions from both clinicians and the AI model. 
The contributions of this paper are:
\begin{itemize}
\item The first human-AI collaborative multi-modal multi-rater learning methodology that produces a highly accurate POD obliteration classifier from T1/T2 MRIs;
\item The first multi-modal multi-rater dataset annotated with imaging and surgery-based POD obliteration labels for the diagnosis of endometriosis.
\end{itemize}
Experiments on our proposed endometriosis dataset shows that our HAICOMM model presents more accurate POD classification than predictions produced by an ensemble of clinicians, by noisy-label learning methods, and by multi-rater learning methods.

\section{Literature Review}
\subsection{Human-AI Collaboration}

Human-AI Collaboration (HAIC) integrates the unique strengths of human experts and AI systems, resulting in improved model capabilities and performance when compared to standalone AI systems~\cite{lu2021human, weitz2019you}. The motivation behind HAIC arises from research~\cite{rosenfeld2018totally, serre2019deep, kamar2012combining} that highlights the limitations of traditional isolated AI methods, overlooking the potential of human-AI collaboration. To overcome these limitations, researchers have proposed various strategies to enhance human-AI collaboration~\cite{bansal2021most, vodrahalli2022humans, humanIntheloop, pradier2021preferential}. Two key strategies within HAIC have emerged: learning to defer and learning to complement. Learning to defer (l2D), which evolved from the concept of learning to reject~\cite{cortes2016learning, madras2018predict}, focuses on optimizing the decision of whether to defer prediction to either the expert or the AI system. Researchers have investigated several L2D approaches~\cite{narasimhan2022post, raghu2019algorithmic, okati2021differentiable}, initially in single-expert scenarios but later extending to multi-user collaborations~\cite{verma2022calibration, mao2023two, ijcai2022-344}. On the other hand, learning to complement~\cite{complement_wilder} focuses on maximizing the expected utility of combined human-AI decisions, and various frameworks have been proposed to model human-AI complementarity~\cite{steyvers2022bayesian, kerrigan2021combining, zhang2023learning, bansal2021most, liu2023humans}.

\subsection{Multi-modal Learning}

Multi-modal learning has become increasingly crucial in various fields, including medical image analysis and computer vision. It combines data from different sources to provide a more comprehensive understanding of tasks. In medical image analysis, several innovative methods have been developed. These include a chilopod-shaped architecture using modality-dependent feature normalization and knowledge distillation~\cite{Dou2020unpaired}, a pixel-wise coherence approach modeling aleatoric uncertainty~\cite{Monteiro2020stochastic}, a trusted multi-view classifier using the Dirichlet distribution~\cite{Han2021trusted}, and an uncertainty-aware model based on cross-modal random network prediction~\cite{Wang2022uncertainty}. Wang et al.~\cite{wang2024enhancing,wang2023learnable,wang2023multi} also tried to approach the missing modality issues in the multi-modal learning scenario. Computer vision has seen advancements in multi-modal learning as well. Researchers have combined channel exchanging with multi-modal learning~\cite{Wang2020deep}, applied self-supervised learning to improve performance~\cite{Patrick2020multi, Patrick2021space}, enhanced video-and-sound source localization~\cite{Chen2021localizing}, introduced a model for multi-view learning~\cite{Jia2020semi}, and explored feature disentanglement methods~\cite{Lee2018diverse, Liu2022learning}.

\subsection{Multi-rater Learning}

Multi-rater learning is a technique designed to train a classifier using noisy labels gathered from multiple annotators. The challenge lies in how to derive a ``clean'' label from these imperfect labels. Traditional approaches often rely on majority voting~\cite{zhou2012ensemble} and the expectation-maximization (EM) algorithm~\cite{rodrigues2014gaussian, raykar2010learning}. Rodrigues et al. ~\cite{rodrigues2018deep} introduce an end-to-end deep neural network (DNN) that incorporates a crowd layer to model the annotator-specific transition matrix, enabling the direct training of a DNN with crowdsourced labels. Alternatively, Chen et al.~\cite{chen2021structured} suggest a probabilistic model that learns an interpretable transition matrix unique to each annotator. Meanwhile, Guan et al.~\cite{guan2018said} employ multiple output layers in the classifier and learn combination weights to aggregate the results. More recently, CROWDLAB~\cite{crowdlab} has set the state of the art in multi-rater learning by using multiple noisy-label samples and predictions with a model trained via label noise learning. Despite the promise of multi-rater learning in leveraging multiple noisy labels per training sample, it falls short by overlooking the concept of human-AI collaboration and multi-modal learning.

\subsection{Imaging-based Endometriosis Detection}

One crucial indicator to detect endometriosis is the obliteration of the Pouch of Douglas (POD)~\cite{kinkel2006diagnosis,Butler2023TheEO}. However, the development of an AI model that can classify such indicator hinges on the availability of precise POD obliteration annotations from T1/T2 MRIs, a task that is challenging because even experienced clinicians often face uncertainty in identifying this sign. Despite these challenges, there have been some efforts to train multi-modal MRI AI models for POD obliteration classification. For example, Zhang et al.~\cite{Zhang_2023} proposed a method to transfer knowledge from ultrasound to MRI for classifying POD obliteration, and Butler et al.~\cite{Butler2023TheEO} explored self-supervised pre-training for multi-modal POD obliteration classification. However, these methods have not been validated against ground-truth labels obtained from surgical reports, making it difficult to 
assess their clinical validity.

Nevertheless, for all of the aforementioned related work, none of the methods deal with human-AI collaboration, multi-modal classification, and multi-rater learning simultaneously, particularly for classifying endometriosis. 
In this paper, we propose the HAICOMM model to address this research gap.

\section{Methodology}

\begin{figure}[t]
\centering
\includegraphics[width=1.0\textwidth]{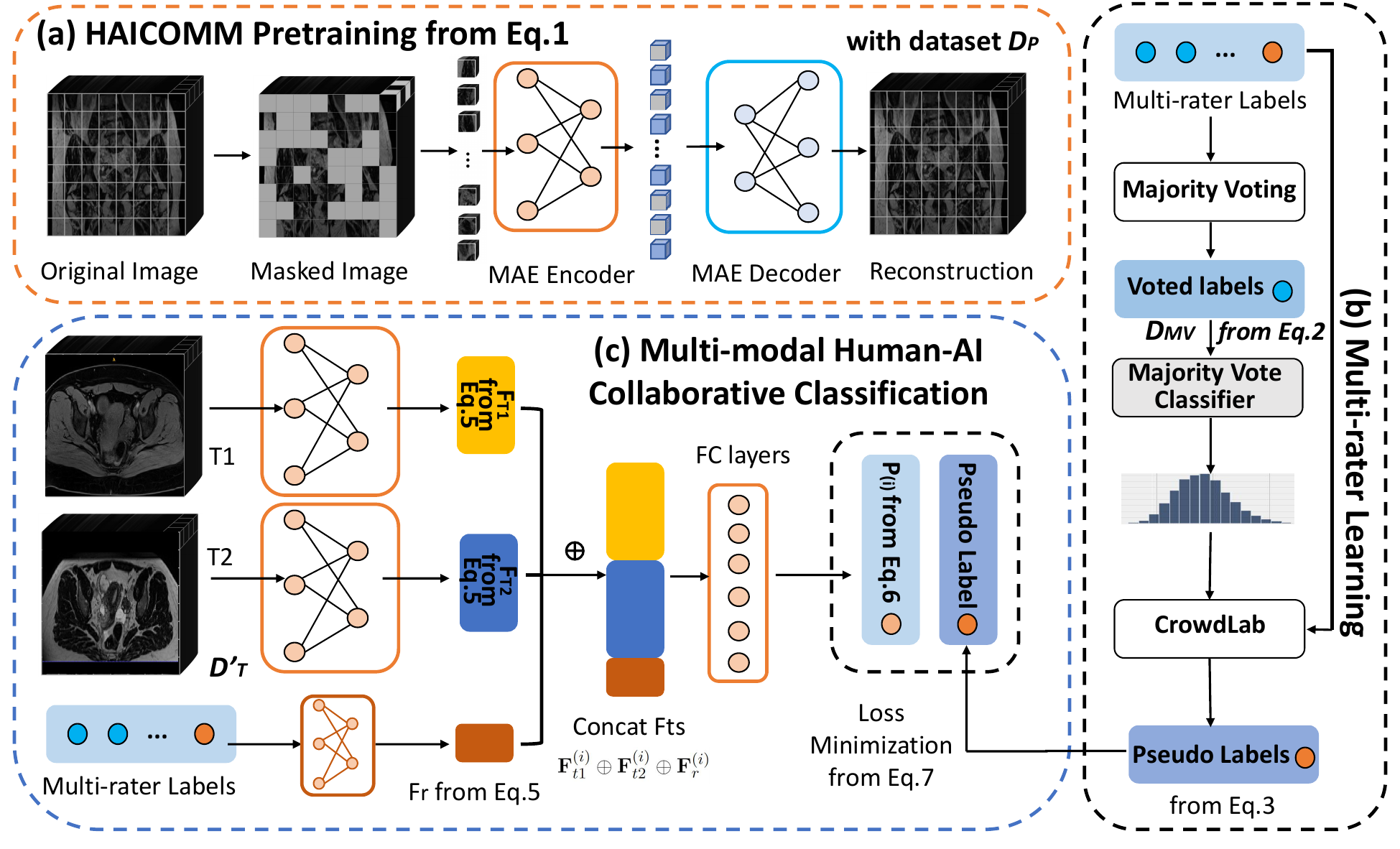}
\caption{\revision{The framework of HAICOMM. The MRI encoders of HAICOMM are: (a) firstly pre-trained with a Masked Autoencoder (MAE) model; then (b) the pseudo clean labels are estimated from the multi-rater learning process; next, (c) the T1 and T2 data, along with the human-produced multi-rater labels are entered into respective feature extraction encoders -- the features from three sources are fused for the final prediction. In the figure, ``FC'' means fully-connected, ``Fts'' represents features, ``Concat'' denotes concatenation, and ``$\oplus$''  is the concatenation operation.}
}
\label{fig:framework}
\end{figure}

The training of our HAICOMM methodology is depicted in Fig.~\ref{fig:framework}.
The first stage consists of pre-training a multi-modal encoder using a large unlabelled T1/T2 MRI dataset, with a self-supervised learning mechanism~\cite{he2022masked} (see frame (a) in Fig.~\ref{fig:framework}).  
Subsequently, for training the proposed human-AI classifier HAICOMM, we first need to estimate the pseudo ground truth label from the the multiple ``noisy'' labels available for each pair of T1/T2 MRI training images. We rely on CrowdLab~\cite{goh2022CrowdLab} to produce such pseudo ground truth labels (see frame (b) in Fig.~\ref{fig:framework}). 
Next, the T1/T2 MRI images with multi-rater (manual) labels are fed into their multi-modal encoders. 
The embeddings from the multi-modal and label encoders are combined to produce the final prediction that is trained to match the pseudo ground truth label (see frame (c) in Fig.~\ref{fig:framework}). 
We provide details about each of these training stages below.

\subsection{Multi-modal Encoder Pre-training}
\label{sec:mae_pretraining}

The MRI encoder of the HAICOMM model is pre-trained with the Masked Autoencoder (MAE) self-supervised learning method~\cite{he2022masked}. For this pre-training, we use a dataset denoted as $ \mathcal{D}_P = \{ \mathbf{x'}_{t1}^{(i)} \}_{i=1}^{M_{t1}} \bigcup \{ \mathbf{x'}_{t2}^{(i)} \}_{i=1}^{M_{t2}}$, with $\mathbf{x'}_{t1}^{(i)},\mathbf{x'}_{t2}^{(i)} \in \mathcal{X} \subset \mathbb{R}^{H \times W \times D}$ denoting the T1 and T2 MRI volumes of size $H \times W \times D$. It is worth noting that the number of unlabeled images, $M = M_{t1}+M_{t2}$, far exceeds the number of labeled images, denoted as $N$ (i.e., $M>>N$), of the datasets that will be defined in Sections~\ref{sec:pseudo clean label Generation} and~\ref{sec:Main Task Training}.

Following the 3D Vision Transformer~\cite{dosovitskiy2020image}, the architecture of 3D-MAE follows an asymmetric encoder-decoder setup. The encoder, parameterized by $\phi$, is represented by $g_\phi:\mathcal{X} \to \mathcal{F}$, which receives visible patches along with positional embeddings that are processed through a 3D Vision Transformer to produce features in the  space $\mathcal{F}$. 
The resulting features are subsequently directed to the decoder, parameterized by $\psi$ and denoted by $f_\psi:\mathcal{F} \to \mathcal{X}$, which reconstructs the original volume with the masked volume tokens. In the MRI pre-training, our objective is to minimize the mean squared error (MSE) of the reconstruction of the original masked patches. Formally, we have:
\begin{equation} \small
\scalebox{0.85}{$
\phi^*, \psi^* = \arg \min _{\phi, \psi} \frac{1}{M} \left ( \sum_{i=1}^{M_{t1}} \Big\| f_\psi\left(g_\phi \left(\mathbf{x'}^{(i)}_{t1}\right) \right) - \mathbf{x'}^{(i)}_{t1} \Big\|^2_2+
\sum_{i=1}^{M_{t2}} \Big\| f_\psi\left(g_\phi \left(\mathbf{x'}^{(i)}_{t2}\right) \right) - \mathbf{x'}^{(i)}_{t2} \Big\|^2_2 \right ),
$}
\label{eq:pretraining}
\end{equation}
\revision{where $\|\cdot\|_2$ denotes the L2-norm.
The optimization in~\eqref{eq:pretraining} jointly learns the encoder parameters $\phi$ and the decoder parameters $\psi$ to minimize the reconstruction error of the masked patches across all training samples. 
For the subsequent training and evaluation of the human-AI collaborative classifier, we adopt the pre-trained feature extractor $g_{\phi^*}(.)$, as explained below in Sec.~\ref{sec:Main Task Training}.
}

\subsection{Multi-rater Learning}
\label{sec:pseudo clean label Generation}

The training of our human-AI collaborative classifier requires each pair of T1/T2 MRI training images to have a single pseudo clean label estimated from the multiple ``noisy'' training labels.
The multi-modal multi-rater dataset is denoted by 
$\mathcal{D}_T=\{ (\mathbf{x}_{t1}^{(i)}, \mathbf{x}_{t2}^{(i)}, \mathbf{y}^{(i)} ) \}_{i=1}^{N}$ with $N$ samples, where the multi-rater label has $K$   binary annotations denoted as $\mathbf{y}^{(i)} \in \mathcal{Y} \subset \{0,1\}^K$, provided by the $K$ clinicians who annotated the training images in $\mathcal{D}_T$.

\revision{With the multi-rater labels, we first perform majority vote to fetch the most frequently appearing labels per training sample. Let us present the majority vote operation as $h:\mathcal{Y} \to \{0,1\}$. Then, we have the mapping from multi-rater labels to the majority label for each multi-modal sample, 
forming the following majority voting dataset:}
\begin{equation} \small
\mathcal{D}_{MV} = \{(\mathbf{x}_{t1}^{(i)}, \mathbf{x}_{t2}^{(i)}, \hat{y}^{(i)} ) \}_{i=1}^{N}, \text{ where } \hat{y}^{(i)} = h(\mathbf{y}^{(i)}),
\end{equation}
\revision{where $\hat{y}^{(i)} \in \{0,1\}$.
With such generated consensus labels from majority vote, we train a classifier $f_{\theta}:\mathcal{X} \times \mathcal{X} \to [0,1]$ that takes the T1 and T2 MRI volumes to optimize a standard binary cross-entropy objective function. 
This trained classifier, together with $\mathcal{D}_T$ will then be used for the filtering process of multi-rater cleaning technique CrowdLab~\cite{goh2022CrowdLab}, denoted by $z:[0,1] \times \mathcal{Y} \to \{0,1\}$, which generates a pseudo clean label for each sample. Formally, CrowdLab's pseudo labeling process is defined by:}
\begin{equation} \small
y^{(i)} = z(f_{\theta}(\mathbf{x}^{(i)}_{t1},\mathbf{x}^{(i)}_{t2}), \mathbf{y}^{(i)}),
\label{eq:generated_pseudo_clean_label}
\end{equation}
where $y^{(i)} \in \{0,1\}$ denotes CrowdLab's estimate for the ``clean'' label of the $i$-th sample, which is referred to as the pseudo ground-truth label.

\subsection{Multi-modal Human-AI Collaborative Classification}
\label{sec:Main Task Training}

\revision{Given the pseudo clean labels $y^{(i)}$ from~\eqref{eq:generated_pseudo_clean_label}, the dataset for the final model training is defined as:}
\begin{equation} \small
\mathcal{D'}_{T} = \{(\mathbf{x}_{t1}^{(i)}, \mathbf{x}_{t2}^{(i)}, \mathbf{y}^{(i)}, y^{(i)} \}_{i=1}^{N}.
\end{equation}

\revision{The pre-trained encoder from~\eqref{eq:pretraining}, parameterized by $\phi^*$, is utilized to initialize the T1 and T2 MRI feature extractors, respectively defined by $g_{\hat{\phi}}:\mathcal{X} \to \mathcal{F}$ and $g_{\phi'}:\mathcal{X} \to \mathcal{F}$, for the classification task. We also need to define a new feature extractor for processing the manual labels with a learnable module defined as $g_{\gamma}:\mathcal{Y} \to \mathcal{F}$. Such manual labels are used in the human-AI collaborative module. Hence, these extractors produce the T1, T2 and rater features as:}
\begin{equation}
    \begin{split}
        \mathbf{F}_{t1}^{(i)} & = g_{\hat{\phi}} (\mathbf{x}_{t1}^{(i)}), \\ 
    \mathbf{F}_{t2}^{(i)} & = g_{\phi'}(\mathbf{x}_{t2}^{(i)}), \\ \mathbf{F}_r^{(i)} &= g_{\gamma}(\mathbf{y}^{(i)}).
    \end{split}
    \label{eqn:train_fts}
\end{equation}
These three feature maps are then concatenated and fed into a learnable linear projection $\pi_\eta:\mathcal{F} \times \mathcal{F} \times \mathcal{F} \to \mathcal{F}$, parameterized by $\eta$ to predict:
\begin{equation} \small
\label{eqn:concat_fts}
    \mathbf{p}^{(i)} = \sigma \left (\pi_\eta(\mathbf{F}_{t1}^{(i)} \oplus \mathbf{F}_{t2}^{(i)} \oplus \mathbf{F}_r^{(i)}) \right),    
\end{equation}
where $\mathbf{p}^{(i)} \in [0,1]^2$ denotes the probabilistic prediction, $\oplus$ represents the concatenation operator, and $\sigma$ is the softmax function.

Finally, the training of the whole model is performed by minimizing the binary cross-entropy loss, as follows:
\begin{equation} \small
\hat{\phi}^*,\phi'^*,\gamma^*,\eta^* = \arg\min_{\hat{\phi},\phi',\gamma,\eta} - \frac{1}{N} \left ( \sum_{i=1}^{N} y^{(i)} \log(p^{(i)}) + (1-y^{(i)}) \log(1-p^{(i)}) \right ),
\end{equation}
\revision{where the $p^{(i)}$ is the predicted probability of the positive class of $i$-th sample (i.e., the $2^{nd}$ dimension $\mathbf{p}^{(i)}$ in~\eqref{eqn:concat_fts}), and $y^{(i)}$ is the pseudo clean label in $\mathcal{D}'_T$. The goal is to learn the optimal parameters $\hat{\phi}^*, \phi'^*, \gamma^*$, and $\eta^*$ that minimize this loss across all $N$ samples.}

The testing process consists of taking the input T1/T2 MRI images and labels $\mathbf{y}$ from clinicians to output $\mathbf{p}$ from~\eqref{eqn:concat_fts}.

\section{\revision{Experimental Settings}}

\subsection{Endometriosis Dataset}

We first introduce our multi-modal multi-rater dataset annotated with imaging and surgery-based POD obliteration labels for the diagnosis of endometriosis. For the pre-training stage, we collected 5,867 unlabeled T1 MRI volumes and 8,984 unlabeled T2 MRI volumes from patients aged between 18 and 45 years old, where the volumes show female pelvis scans obtained from various MRI machines with varying resolutions.
\revision{The pre-training pelvic MRI scans were obtained from the South Australian Medical Imaging (SAMI) service. For this paper, we included only pelvic MRI scans of female patients acquired using either T1- or T2-weighted imaging sequences. These scans were collected from multiple centers using various equipments, including the 1.5T Siemens Aera, 3T Siemens TrioTim, 1.5T Philips Achieva, 1.5T Philips Ingenia, and 1.5T Philips Intera. The T1 and T2 datasets include a wide variety of sequences, where each scan is acquired using a combination of the following parameters: spin echo sequences (e.g., Turbo Spin Echo, TSE) or gradient echo sequences (e.g., Volumetric Interpolated Breath-hold Examination, VIBE); orientation in either 2D planes (axial, sagittal, coronal) or 3D sequences (e.g., SPC); with or without fat saturation/suppression techniques (e.g., DIXON, SPIR, SPAIR); and with or without gadolinium contrast enhancement. We show the statistics of the size distribution
information of the pre-training dataset in Table~\ref{tab:combined_subtables_pretraining}.}
\revision{As described in Section \ref{sec:mae_pretraining}, this dataset is used as containing 14,851 independent scans for the Masked Autoencoder (MAE) self-supervised learning. To standardize the data, the volumes were resampled to $1\text{mm} \times 1\text{mm} \times 3\text{mm}$ voxels. Also, 
we apply 3D contrast-limited adaptive histogram equalization (CLAHE) to enhance image local contrast and refine edge definitions. We choose CLAHE, because it produced the best classification results among the different data pre-processing methods tested on an internal validation process (e.g., voxel value truncate.).
}

\begin{table}[ht] \small
    \centering
    \caption{\revision{Statistics of the volume size distribution  of the pre-training dataset.}}
    \begin{minipage}{0.45\textwidth}
        \centering
        \subcaption{T1 pre-training dataset}
        \begin{tabular}{c|c|c|c|c}
            \hline \hline
            Dim & Mean & Std & Max & Min \\
            \hline
            0 & 326.67 & 93.73 & 550 & 66 \\
            1 & 271.97 & 88.78 & 530 & 60 \\
            2 & 79.64 & 24.5 & 430 & 50 \\
            \hline \hline
        \end{tabular}
    \end{minipage}
    \hspace{0.05\textwidth}
    \begin{minipage}{0.45\textwidth}
        \centering
        \subcaption{T2 pre-training dataset}
        \begin{tabular}{c|c|c|c|c}
            \hline \hline
            Dim & Mean & Std & Max & Min \\
            \hline
            0 & 245.78 & 113.33 & 530 & 60 \\
            1 & 208.47 & 93.87 & 504 & 50 \\
            2 & 77.01 & 23.99 & 430 & 50 \\
            \hline \hline
        \end{tabular}
    \end{minipage}
    \label{tab:combined_subtables_pretraining}
\end{table}

\revision{For the training of the human-AI collaborative POD obliteration classifier, we collected 
82 pairs of T1/T2 MRI volumes with patients aged 18 to 45 years old. 
These scans were obtained across multiple clinical sites, with each case annotated by three experienced clinicians who work in  clinics specialized in the imaging-based diagnosis of endometriosis. 
The scans were acquired using a standardized protocol for endometriosis examination and they show a specific region surrounding the uterus, which is the area where the signs of POD obliteration are more visible. 
MRIs from patients with a history of hysterectomy or those with large pelvic lesions that limited adequate assessment of the POD were excluded. 
The scanners are of the following models: SIEMENS Aera, SIEMENS Espree, SIEMENS MAGNETOM Sola, and SIEMENS MAGNETOM Vida. 
The volume types are: T1-weighted MRI images, obtained using a Volume Interpolated Breath-hold Examination technique with Dixon fat-water separation, acquired in the transverse plane; and T2-weighted MRI images, acquired using the SPACE sequence, characterized by non-saturation techniques, taken in the coronal plane, and with isotropic voxel dimensions.
All 3D volumes were reoriented to a Right-Anterior-Superior (RAS) coordinate system to standardize anatomical alignment, then resampled with an output spacing. Similar to the pre-training, to standardize the data, the volumes were resampled to $1\text{mm} \times 1\text{mm} \times 3\text{mm}$ voxels. We show the statistics of the size distribution
information of the POD obliteration dataset in Table~\ref{tab:combined_subtables_training}}

\begin{table}[ht] \small
    \centering
    \caption{\revision{Statistics of the volume size distribution of the POD obliteration dataset.}}
    \begin{minipage}{0.45\textwidth}
        \centering
        \subcaption{T1 modality in the dataset}
        \begin{tabular}{c|c|c|c|c}
            \hline \hline
            Dim & Mean & Std & Max & Min \\
            \hline
            Depth & 383.91 & 91.88 & 576 & 260 \\
            Height & 312.10 & 72.61 & 464 & 250 \\
            Width & 72.16 & 8.46 & 80 & 51 \\
            \hline \hline
        \end{tabular}
    \end{minipage}
    \hspace{0.05\textwidth}
    \begin{minipage}{0.45\textwidth}
        \centering
        \subcaption{T2 modality in the dataset}
        \begin{tabular}{c|c|c|c|c}
            \hline \hline
            Dim & Mean & Std & Max & Min \\
            \hline
            Depth & 389.80 & 18.40 & 448 & 384 \\
            Height & 192.30 & 26.22 & 336 & 160 \\
            Width & 289.71 & 19.37 & 354 & 192 \\
            \hline \hline
        \end{tabular}
    \end{minipage}
    \label{tab:combined_subtables_training}
\end{table}

\revision{This training set with 82 volumes is subdivided into 62 volumes for training and 20 for validation that are used for cross validation for hyper-parameter selection.} 
We further collect 30 cases that contain ground truth annotation of POD obliteration from surgical reports.
These cases serve as gold standards for testing. 
\revision{We show in Table~\ref{tab:abnormal} the distribution of normal  vs abnormal (i.e., POD obliteration) images in the training and testing sets. Note that while in the testing set, we have ground truth from surgery, in the training set, we only have the labelers' annotation.
We also use CLAHE pre-processing in this dataset. Similar to the pre-training, to standardize the data, the volumes were resampled to $1\text{mm} \times 1\text{mm} \times 3\text{mm}$ voxels.}

\begin{table}[h!]
\caption{\revision{Number of normal and abnormal (i.e., with POD obliteration) images in the training and testing sets. In the training set, `A1' to `A3' are the three annotators, `MV' represents the annotations from majority voting, and `CL' denotes the annotations from CrowdLab~\cite{crowdlab}. In the testing set, we have the grount truth (GT) from surgery. 
}}
\label{tab:abnormal}
\begin{center}
\scalebox{0.95}{
\begin{tabular}{c|ccccc|c}
\hline \hline
POD         & \multicolumn{5}{c|}{Training Set}                                                                              & Testing Set \\ \hline
         & \multicolumn{1}{c}{A1} & \multicolumn{1}{c}{A2} & \multicolumn{1}{c}{A3} & \multicolumn{1}{c}{MV} & CL &  GT       \\ \hline
Normal   & \multicolumn{1}{c}{62} & \multicolumn{1}{c}{48} & \multicolumn{1}{c}{36} & \multicolumn{1}{c}{51} & 47 & 15      \\ \hline
Abnormal & \multicolumn{1}{c}{20} & \multicolumn{1}{c}{34} & \multicolumn{1}{c}{46} & \multicolumn{1}{c}{31} & 34 & 15      \\ \hline \hline
\end{tabular}
}\end{center}
\end{table}

\subsection{Implementation Details}

\revision{For model pre-training, the input volumes are cropped and possibly zero-padded to achieve the dimension of 64 × 128 × 128 voxels. To maintain consistency with the pre-training dataset, the endometriosis training and testing samples are manually centered at the uterine region which is the most important region for the detection of POD obliteration. Later we center crop each volume to the same dimensions as pre-training data, i.e. $64 \times 128 \times 128$ voxels, for training and testing.
In both pre-training and the training of the human-AI collaborative POD obliteration classifier, the multi-modal encoder for each modality is a transformer with 12 blocks. 
The majority vote classifier has a 3D-ResNet50 as its backbone network~\cite{hara2017learning}. 
For the human-AI collaborative POD obliteration classifier training, we use 5 epochs for model optimization warming up. 
AdamW optimizer and base learning rate of 1e-3 with cosine annealing \cite{loshchilov2016sgdr} learning rate tuning strategy are adopted. 
Three multi-rater labels from three different annotators are incorporated into the training process. 
In the testing phase, the scans are also cropped in the uterine regions and the clinical surgical results serve as the ground truth for accurate evaluation. 
Using a cross-validation procedure, with a training set containing 62 volumes and a validation set comprising 20 volumes, we observe signs of overfitting after 60 epochs. To prevent this, we implemented early stopping and halted the training at the 60th epoch. 
Note that the majority voting is only produced for the consensus pseudo label required for the training of the model $f_{\theta}(.)$ to be used by CrowdLab, as explained in Eq.~\ref{eq:generated_pseudo_clean_label}. Once the pseudo clean labels are generated by CrowdLab, the majority voting will no longer be needed. 
All experiments in the paper are run on an NVIDIA GeForce RTX 3090. The main Python libraries used in our implementation are: torch 1.7.1+cu110, torchvision 0.8.2+cu110, scikit-image and scipy.}

\subsection{Quantitative Evaluation Settings}

We compare the performance of our proposed HAICOMM with respect to the following models: 1) purely manual annotation from the three expert clinicians via majority voting;
2) models trained with noisy-label learning techniques (SSR~\cite{Feng2022SSR} and ProMix~\cite{xiao2023promix}) using the noisy labels from one of the annotators (GT1, GT2, GT3);
3) models trained from labels produced by the multi-rater learning CrowdLab~\cite{goh2022CrowdLab} (in the table denoted as models w/ CL);
and 
4) human-AI classifiers using the three annotators (models w/ HAIC).
In terms of evaluation metrics, we adopt Accuracy and Area Under the ROC Curve (AUROC).

\revision{We also show an ablation study that compares the performance of each annotator without using the system, with the performance of different combinations of annotators to be used in the human-AI collaboration, and the performance of the proposed HAICOMM.
}

\section{\revision{Results and Discussion}}
\subsection{\revision{Model Performance}}

\begin{table}[t!]
\caption{Performance comparison of HAICOMM and other models \revision{on the test set in terms of accuracy and AUROC (and respective improvements of HAICOMM) with the testing ground-truth labels from surgical reports}.
\revision{The standard deviation is calculated by inference time bootstrapping.}
``Majority Vote'' denotes a purely manual classification using the majority vote of the three annotators. SSR and ProMix w/ ``GT1'', ``GT2'' and ``GT3'' mean noisy-label learning models trained with labels from annotators \#1, \#2 and \#3, respectively. ``CL'' denotes noisy-label learning models trained with labels $y^{(i)}$ from Eq.~\eqref{eq:generated_pseudo_clean_label} produced by the multi-rater learning method CrowdLab. 
``HAIC'' represents a model that collaborates with the annotators and that is trained with labels from CrowdLab.
The best results for each column are in bold.
}
\label{tab:overall}
\begin{center}
\scalebox{0.85}{
\begin{tabular}{l|l|cc|cc}
\hline \hline
Methods &Models          & Accuracy     & Improvement & AUROC           & Improvement \\  
\hline
Human    & Majority Vote  & 0.70±0.00                         & 14.29\%    & -                      & -                    \\ \hline
         & SSR w/ GT1     & 0.58±0.02                         & 37.93\%    & 0.58±0.01          &   53.83\%              \\
Noisy    & SSR w/ GT2     & 0.60±0.04 & 33.33\%    & 0.58±0.01          & 51.67\%              \\
Label    & SSR w/ GT3     & 0.57±0.02                         & 41.17\%    & 0.56±0.06          & 57.07\%              \\
Learning & ProMix w/ GT1  & 0.53±0.02                         & 50.01\%    & 0.54±0.04          & 64.75\%              \\
         & ProMix w/ GT2  & 0.62±0.03                         & 29.72\%    & 0.58±0.02          & 52.63\%              \\
         & ProMix w/ GT3  & 0.52±0.05                         & 54.83\%    & 0.57±0.02          & 56.54\%              \\ \hline
Multi-   & SSR w/ CL      & 0.62±0.02                         & 29.72\%    & 0.59±0.01          & 50.82\%              \\
rater    & ProMix w/ CL   & 0.65±0.04                         & 23.08\%    & 0.54±0.03          & 63.32\%              \\ \hline
HAIC     & SSR w/ HAIC    & 0.68±0.08                         & 17.08\%    & 0.74±0.01          & 19.37\%              \\
         & ProMix w/ HAIC & 0.67±0.06                         & 20.00\%    & 0.74±0.04          & 19.98\%              \\ \hline
Ours     & HAICOMM        & \textbf{0.80±0.04}                & \textbf{-} & \textbf{0.89±0.06} & -                    \\ \hline \hline
\end{tabular}
}\end{center}
\end{table}

\begin{figure}[t]
\centering
\includegraphics[width=0.6\textwidth]{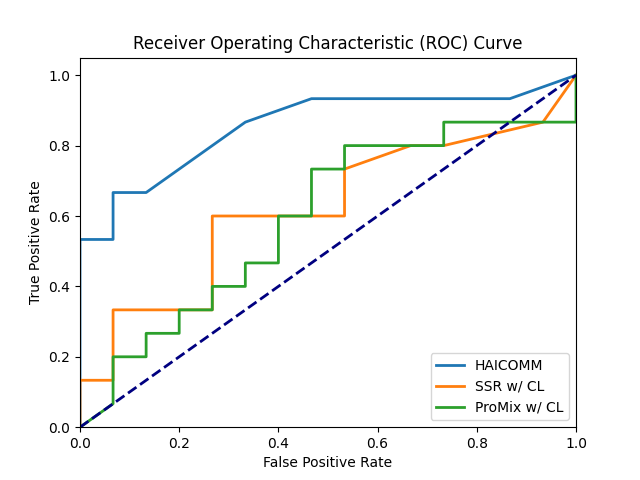}
\caption{\revision{The ROC Curve of HAICOMM models and its counterparts.}}
\label{fig:roc}
\end{figure}

The performance results in Table~\ref{tab:overall} show that the proposed HAICOMM outperforms other competing models by a large margin across the accuracy and AUC measures. 
Relative improvements vary from 9.10\% to 54.83\% on accuracy and 19.37\% and 64.75\% on AUROC. 
The standard deviation is calculated by inference time bootstrapping. 
\revision{Also, we provide the the ROC Curve of HAICOMM models and its counterparts SSR and ProMix with CrowdLab labels in Fig.~\ref{fig:roc}.}

There are interesting points to observe in the results from Table~\ref{tab:overall}. 
First, the multi-rater learning tends to be more accurate than noisy label learning.
The manual annotation without any assistance from the model $f_{\theta}(.)$ in Eq.~\ref{eq:generated_pseudo_clean_label}, shows a relatively low accuracy of 0.7, motivating the importance of the proposed human-AI collaboration.
Also, when noisy-label learning models are designed to collaborate with humans, we can see large performance improvements, such as shown by ``SSR w/ HAIC'' and ``ProMix w/ HAIC''. 
However, the proposed HAICOMM still obtains much higher accuracy and AUROC. 
Additionally, the proposed HAICOMM shows a much simpler training algorithm than ``SSR w/ HAIC'' and ``ProMix w/ HAIC''. 
To summarize, the proposed model  outperforms not only the ensemble of experts (Majority Vote), but also the top-performing multi-rater learning model (SSR w/CL and ProMix w/ CL), as well as the best noisy-label learning methods (SSR and ProMix), even after adding human AI collaboration (SSR w/ HAIC and ProMix w/ HAIC) by a large margin.

\subsection{\revision{Human-AI Collaborative Multi-modal Multi-rater Ablation Study}}

\begin{table}[t!]
\caption{Accuracy and AUROC performance analyses of HAICOMM and its variants \revision{on the test set}. ``R1'', ``R2'' and ``R3'' denote models trained with input annotations from annotators \#1, \#2 and \#3, respectively. ``HAIC'' represents model trained with multi-rater labels inputs for human-AI collaborations. T1 and T2 Only w/ HAIC represent single-modality HAICOMM approaches.
The best results for each column are in bold.
}
\label{tab:analyses}
\begin{center}
\scalebox{0.8}{
\begin{tabular}{l|c|c}
\hline \hline
Models           & Accuracy & AUROC  \\ \hline
Labels from Rater \#1 (R1)     & 0.67   & - \\
Labels from Rater \#2 (R2)    & 0.73   & - \\
Labels from Rater \#3 (R3)    & 0.70   & - \\ \hline
HAICOMM w/o HAIC \ \ & 0.63   & 0.59 \\
HAICOMM w/ R1     & 0.77   & 0.70 \\
HAICOMM w/ R2     & \textbf{0.80}   & 0.84 \\
HAICOMM w/ R3     & 0.57   & 0.60 \\
HAICOMM w/ R1,2   & 0.77   & 0.87 \\
HAICOMM w/ R2,3   & 0.73   & 0.87      \\
HAICOMM w/ R1,3   & 0.63   & 0.67  \\ \hline
T1 Only w/ HAIC   & 0.67   & 0.81 \\
T2 Only w/ HAIC   & 0.77   & 0.88 \\ \hline
HAICOMM          & \textbf{0.80}   & \textbf{0.89} \\ \hline \hline
\end{tabular}
}\end{center}
\end{table}

The first three rows of Table~\ref{tab:analyses} present the accuracy of each of the three annotators.
The next rows show HAICOMM without relying on any human collaboration (w/o HAIC), then the next 6 rows show different combinations of annotators for the human-AI collaboration process. This is followed by two rows showing HAICOMM with single modality data (either T1 or T2) in the input. Last row shows the HAICOMM results. Note that the collaboration with annotators almost always improve over the result of HAICOMM w/o HAIC, and it also improves the accuracy for most of the annotators (particularly R1 and R2).
Interestingly, we found that the model with R2 inputs performs the best among with single rater labels. The model with combination inputs of R1 and R3 performs the worst. 
This may suggest that R2 provides  relatively more accurate labels compared to R1 and R3. This phenomenon resonates with the fact that R2 provides the most accurate labels among three raters (as shown in the first three rows). 
This table also shows that both single modality results with HAIC (with T2 being much better than T1) have worse results than the multi-modal HAICOMM, which provides evidence of the need for multi-modal analysis in the classification of POD obliteration.

\subsection{\revision{Analyses}}

\begin{figure}[h!]
    \centering
    \begin{subfigure}[b]{0.75\textwidth}
        \centering
        \includegraphics[width=\textwidth]{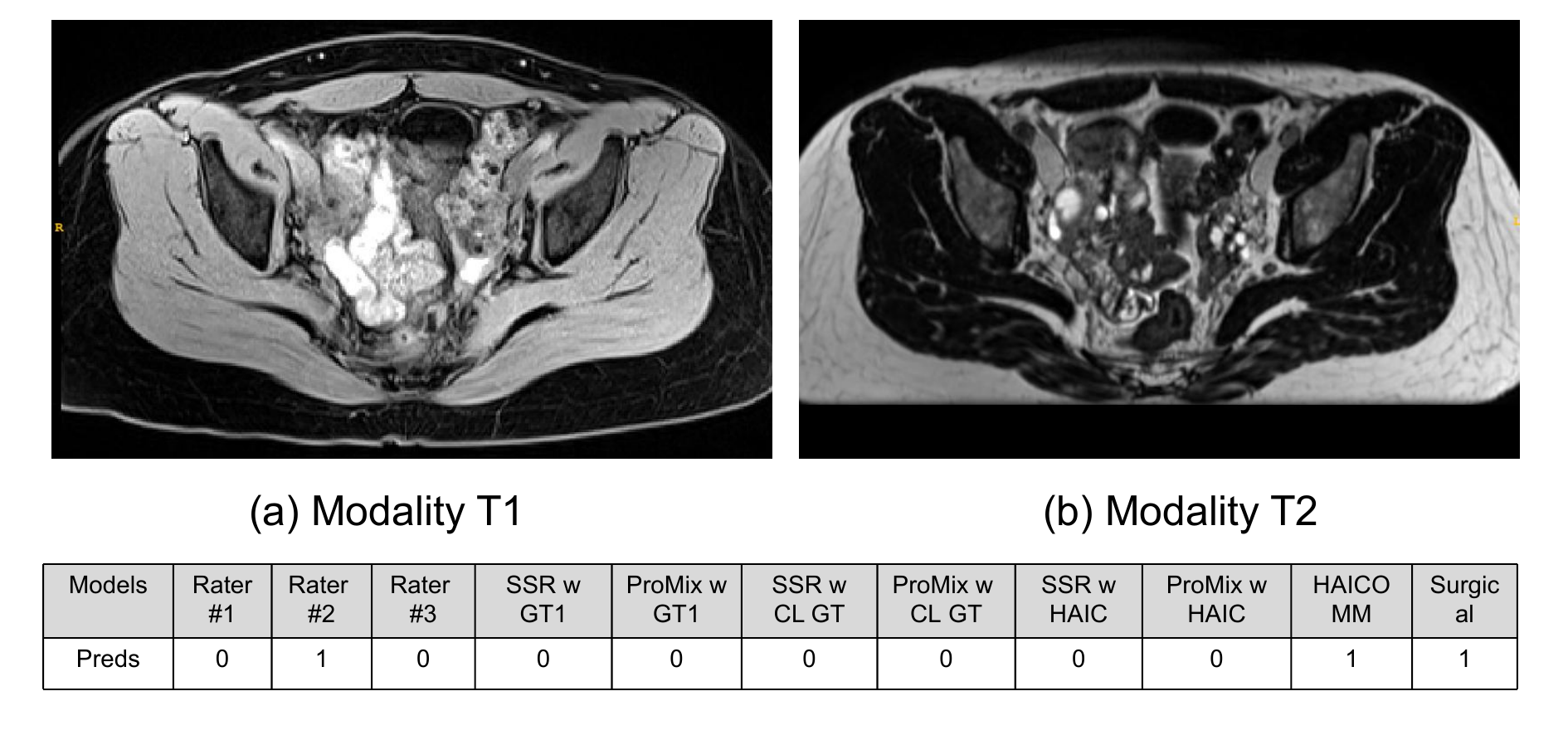}
    \end{subfigure}
    \hfill
    \begin{subfigure}[b]{0.75\textwidth}
        \centering
        \includegraphics[width=\textwidth]{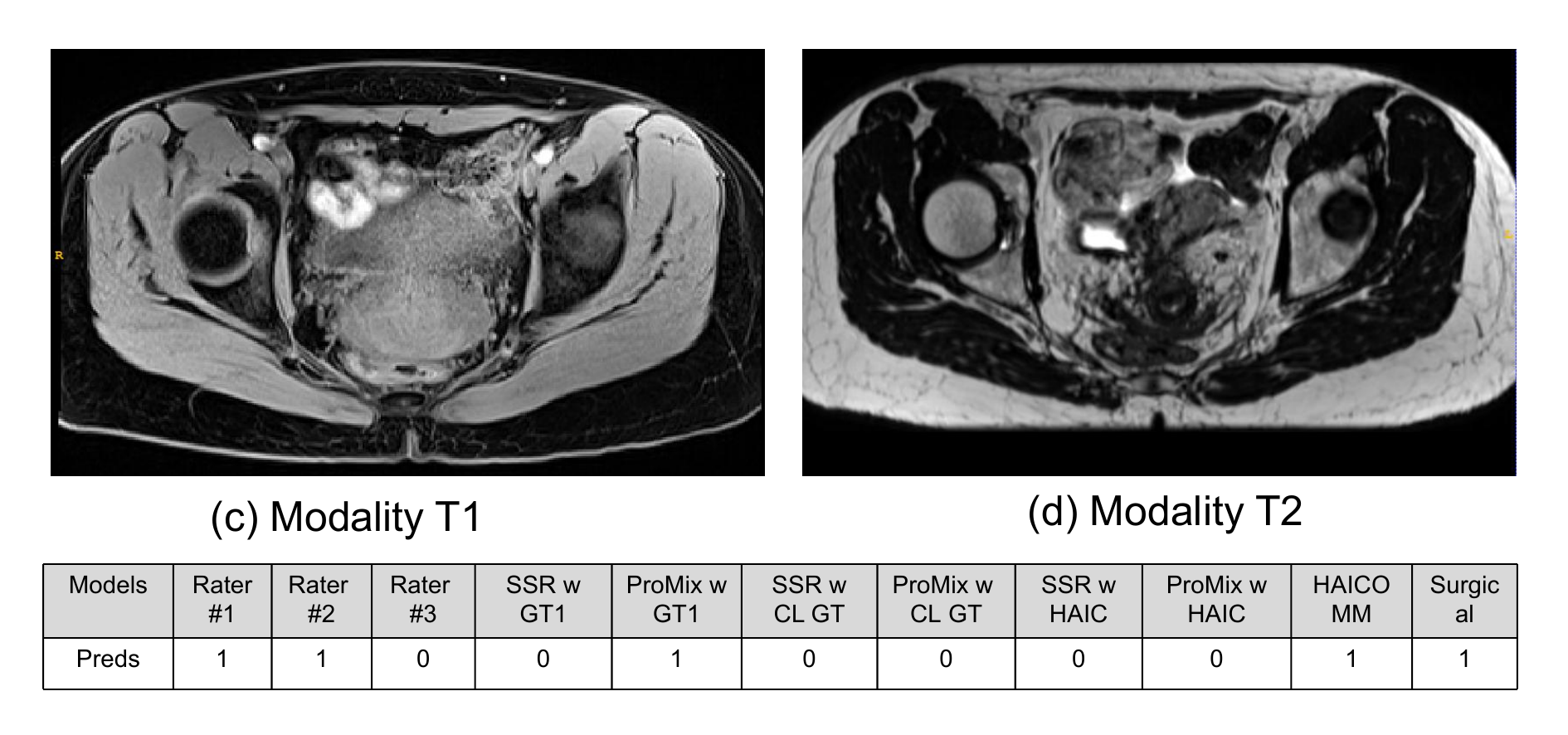}
    \end{subfigure}
\caption{\revision{Qualitative analysis of HAICOMM. Each row (a),(b) and (c),(d) presents the input T1 and T2 MRI images, with corresponding tables below showing predictions from three human raters (Rater \#1, \#2, and \#3). The tables also display predictions from SSR and ProMix models trained using labels from Rater \#1 (SSR w/ GT1, ProMix w/ GT1) and CROWDLAB (SSR w/ CL GT, ProMix w/ CL GT). Following these, we provide the predictions from SSR and ProMix models trained on CROWDLAB labels and utilizing human-AI collaborative classification (SSR w/ HAIC, ProMix w/ HAIC). Finally, we present results from our HAICOMM model, along with the ground truth label based on surgical data.}}
\label{fig:qualitative}
\end{figure}

\revision{We conduct a qualitative analysis of HAICOMM. In Figure~\ref{fig:qualitative}, (a) and (b) are the input T1 and T2 MRIs, respectively. The table below shows the predictions by the three raters (Rater \#1,\#2,\#3), then the predictions by SSR and ProMix trained with Rater \#1's labels and CROWDLAB's labels  (SSR w GT1, ProMix w GT1, SSR w CL GT, ProMix w CL GT). Next, we show SSR and ProMix trained with CROWDLAB's labels and relying on human-AI collaborative classification (SSR w HAIC, ProMix w HAIC), followed by the result from our HAICOMM, and the ground truth label from surgical data. This figure shows that the proposed HAICOMM model can generate correct labels, with other approaches showing incorrect predictions. For (c) and (d), we show that the proposed HAICOMM model produced a correct label while most of other methods failed (only ProMix w GT1 and HAICOMM predict the surgical ground truth label correctly).}

\begin{figure}[h!]
\centering
\includegraphics[width=0.8\textwidth]{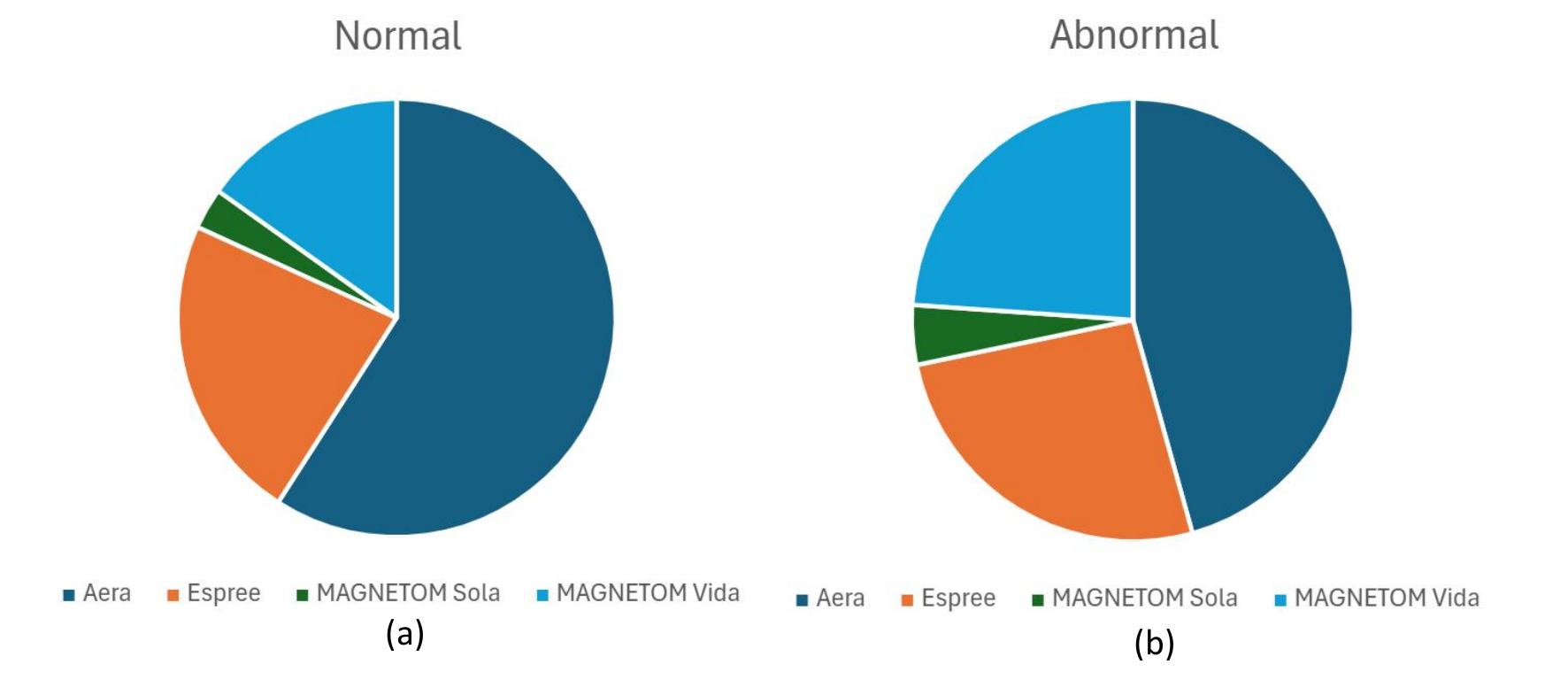}
\caption{\revision{Distribution of manufacturer model of scanners for the normal (a) and abnormal (b)  datasets.}}
\label{fig:train_test_device}
\end{figure}

\begin{figure}[h!]
\centering
\includegraphics[width=0.8\textwidth]{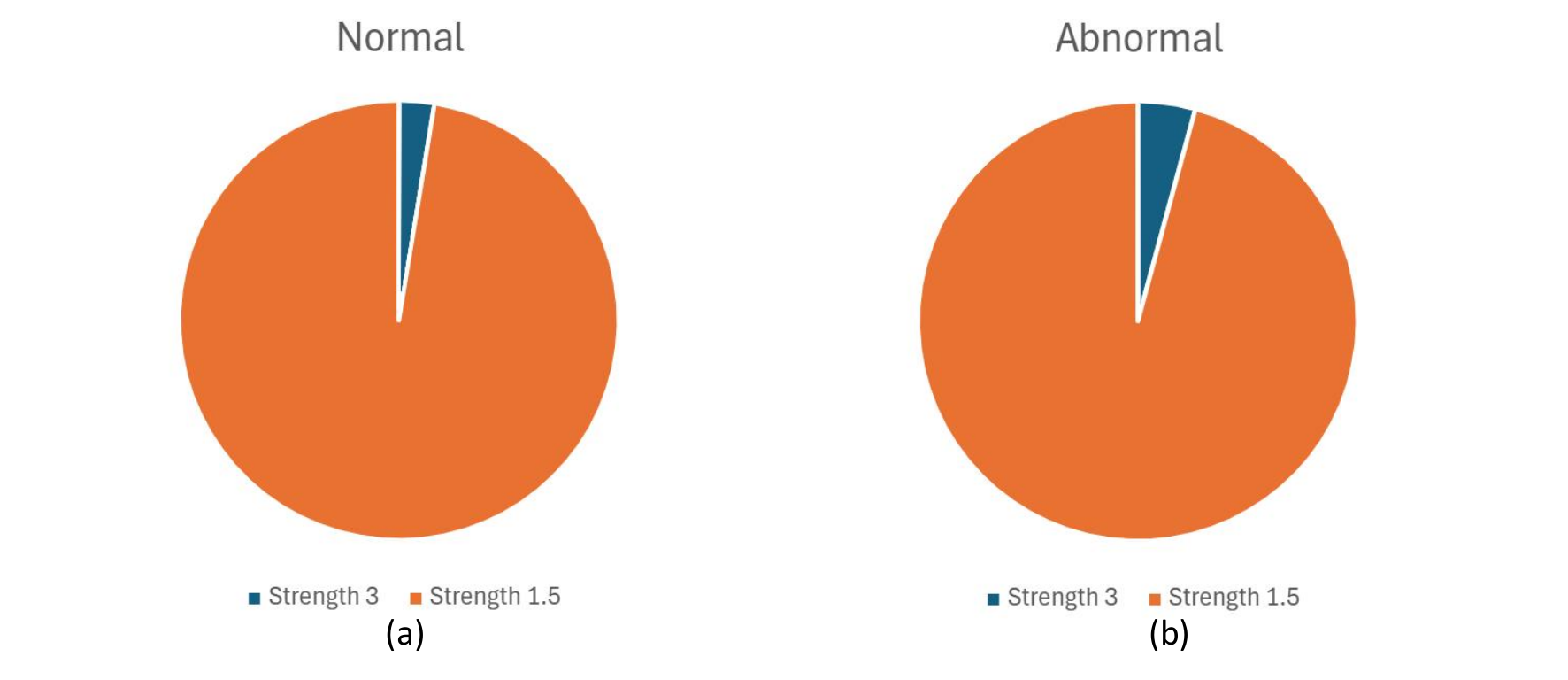}
\caption{\revision{Distribution of magnetic field strength of scanners for the normal (a) and abnormal (b)  datasets.}}
\label{fig:train_test_device2}
\end{figure}

\revision{In terms of the normal and abnormal volumes, they are sampled from distributions with equivalent condition / equipment / resolution, so they have similar properties. We calculate the statistics of the volume size distribution of the normal / abnormal (for the training data, we use majority vote results as the label) ones in Table~\ref{tab:normal_t1} and Table~\ref{tab:normal_t2}. Moreover, we plot the distribution of manufacturer model of scanners for normal and abnormal in Figure~\ref{fig:train_test_device} and the distribution of magnetic field strength for
normal and abnormal in Figure~\ref{fig:train_test_device2}.}

\begin{table}[h!]
\caption{\revision{Number of normal and abnormal (i.e., with POD obliteration) volumes in the training and testing sets. In the training set, `A1' to `A3' are the three annotators, `MV' represents the annotations from majority voting, and `CL' denotes the annotations from CrowdLab~\cite{crowdlab}. In the testing set, we have the grount truth (GT) from surgery.
}}
\label{tab:abnormal}
\begin{center}
\scalebox{0.75}{
\begin{tabular}{c|ccccc|c}
\hline \hline
POD         & \multicolumn{5}{c|}{Training Set}                                                                              & Testing Set \\ \hline
         & \multicolumn{1}{c}{A1} & \multicolumn{1}{c}{A2} & \multicolumn{1}{c}{A3} & \multicolumn{1}{c}{MV} & CL &  GT       \\ \hline
Normal   & \multicolumn{1}{c}{62} & \multicolumn{1}{c}{48} & \multicolumn{1}{c}{36} & \multicolumn{1}{c}{51} & 47 & 15      \\ \hline
Abnormal & \multicolumn{1}{c}{20} & \multicolumn{1}{c}{34} & \multicolumn{1}{c}{46} & \multicolumn{1}{c}{31} & 34 & 15      \\ \hline \hline
\end{tabular}
}\end{center}
\end{table}

\begin{table}[h] \small
    \centering
    \caption{\revision{Statistics of the volume size distribution  of the T1 MRIs of POD obliteration dataset.}}
    \begin{minipage}{0.45\textwidth}
        \centering
        \subcaption{Normal Statistics}
        \begin{tabular}{c|c|c|c|c}
            \hline \hline
            Dim & Mean & Std & Max & Min \\
            \hline
            Depth & 382.73 & 91.05 & 576 & 260 \\
            Height & 313.67 & 74.57 & 464 & 260 \\
            Width & 72.44 & 7.81 & 80 & 52 \\
            \hline \hline
        \end{tabular}
    \end{minipage}
    \hspace{0.05\textwidth}
    \begin{minipage}{0.45\textwidth}
        \centering
        \subcaption{Abnormal Statistics}
        \begin{tabular}{c|c|c|c|c}
            \hline \hline
            Dim & Mean & Std & Max & Min \\
            \hline
            Depth & 381.71 & 90.31 & 576 & 320 \\
            Height & 309.35 & 71.95 & 464 & 250 \\
            Width & 72.07 & 8.67 & 80 & 51 \\
            \hline \hline
        \end{tabular}
    \end{minipage}
    \label{tab:normal_t1}
\end{table}

\begin{table}[h] \small
    \centering
    \caption{\revision{Statistics of the volume size distribution  of the T2 MRIs of POD obliteration dataset.}}
    \begin{minipage}{0.45\textwidth}
        \centering
        \subcaption{Normal Statistics}
        \begin{tabular}{c|c|c|c|c}
            \hline \hline
            Dim & Mean & Std & Max & Min \\
            \hline
            Depth & 389.82 & 18.54 & 448 & 384 \\
            Height & 190.06 & 17.64 & 256 & 176 \\
            Width & 292.36 & 13.90 & 336 & 288 \\
            \hline \hline
        \end{tabular}
    \end{minipage}
    \hspace{0.05\textwidth}
    \begin{minipage}{0.45\textwidth}
        \centering
        \subcaption{Abnormal Statistics}
        \begin{tabular}{c|c|c|c|c}
            \hline \hline
            Dim & Mean & Std & Max & Min \\
            \hline
            Depth & 389.79 & 18.40 & 448 & 384 \\
            Height & 193.05 & 28.50 & 336 & 160 \\
            Width & 288.82 & 20.83 & 354 & 192 \\
            \hline \hline
        \end{tabular}
    \end{minipage}
    \label{tab:normal_t2}
\end{table}

\section{Conclusion and Future Work}

In this paper, we proposed the Human-AI Collaborative Multi-modal Multi-rater Learning (HAICOMM) methodology for an imaging-based endometriosis classification. It integrates the capabilities of machine learning models and multiple human labels to enhance the classification accuracy of POD obliteration from T1/T2 MRIs. Evaluation on our endometriosis dataset demonstrates the efficacy of the HAICOMM model, surpassing ensemble clinician predictions, noisy-label learning approaches, and a multi-rater learning method. This underscores the potential of collaborative efforts between AI and human clinicians in diagnosing and managing endometriosis and other complex medical conditions.
To the best of knowledge, we are the first to propose the multi-modal multi-rater classification task. Furthermore, our endometriosis dataset is the first in the field to enable the development of multi-modal multi-rater classifiers.

One potential limitation of our method is the dataset size. Currently, we are dedicated to collect more data from different clinical sources to expand the dataset. The use of such multiple clinical sources will require the exploration of domain adaptation techniques to enable a better flexibility of the method to work in multiple domains.
Beyond this issue, the need for specific labellers for training and testing is another potential limitation. 
We plan to address this issue with the development of techniques that work with a variable set of labellers during training and testing.
Another interesting direction is the collection of new datasets for other multi-modal multi-rater clinical problems to enable the evaluation of our HAICOMM in a different task.
\revision{One more potential future direction is the development of a detector of region of interests, which will require dense (i.e., pixel-wise) annotations of the training and testing sets and the collection of significantly larger training sets.}

\vspace{0.8cm}

\noindent\textbf{References}
\bibliographystyle{elsarticle-num} 
\bibliography{mybib}

\begin{thebibliography}{10}
\expandafter\ifx\csname url\endcsname\relax
  \def\url#1{\texttt{#1}}\fi
\expandafter\ifx\csname urlprefix\endcsname\relax\def\urlprefix{URL }\fi
\expandafter\ifx\csname href\endcsname\relax
  \def\href#1#2{#2} \def\path#1{#1}\fi

\bibitem{lagana2020evaluation}
A.~S. Lagana, F.~M. Salmeri, H.~Ban~Frange{\v{z}}, F.~Ghezzi, E.~Vrta{\v{c}}nik-Bokal, R.~Granese, Evaluation of m1 and m2 macrophages in ovarian endometriomas from women affected by endometriosis at different stages of the disease, Gynecological Endocrinology 36~(5) (2020) 441--444.

\bibitem{moss2021delayed}
K.~Moss, J.~Doust, H.~Homer, I.~Rowlands, R.~Hockey, G.~Mishra, Delayed diagnosis of endometriosis disadvantages women in art: a retrospective population linked data study, Human Reproduction 36~(12) (2021) 3074--3082.

\bibitem{australian2019endometriosis}
A.~I. of~Health, Welfare, Endometriosis in Australia: Prevalence and Hospitalisations: In Focus, Australian Institute of Health and Welfare, 2019.

\bibitem{becker2022eshre}
C.~M. Becker, et~al., Eshre guideline: endometriosis, Human reproduction open 2022~(2) (2022) hoac009.

\bibitem{horne2022pathophysiology}
A.~W. Horne, S.~A. Missmer, Pathophysiology, diagnosis, and management of endometriosis, bmj 379 (2022).

\bibitem{soliman2016direct}
A.~M. Soliman, H.~Yang, E.~X. Du, C.~Kelley, C.~Winkel, The direct and indirect costs associated with endometriosis: a systematic literature review, Human reproduction 31~(4) (2016) 712--722.

\bibitem{kinkel2006diagnosis}
K.~Kinkel, K.~A. Frei, C.~Balleyguier, C.~Chapron, Diagnosis of endometriosis with imaging: a review, European radiology 16 (2006) 285--298.

\bibitem{Butler2023TheEO}
D.~Butler, H.~Wang, Y.~Zhang, M.-S. To, G.~Condous, M.~Leonardi, S.~Knox, J.~Avery, L.~Hull, G.~Carneiro, \href{https://api.semanticscholar.org/CorpusID:266196804}{The effectiveness of self-supervised pre-training for multi-modal endometriosis classification*†}, 2023 45th Annual International Conference of the IEEE Engineering in Medicine \& Biology Society (EMBC) (2023) 1--5.
\newline\urlprefix\url{https://api.semanticscholar.org/CorpusID:266196804}

\bibitem{kataoka2005posterior}
M.~L. Kataoka, et~al., Posterior cul-de-sac obliteration associated with endometriosis: Mr imaging evaluation, Radiology 234~(3) (2005) 815--823.

\bibitem{indrielle2020diagnostic}
T.~Indrielle-Kelly, et~al., Diagnostic accuracy of ultrasound and mri in the mapping of deep pelvic endometriosis using the international deep endometriosis analysis (idea) consensus, BioMed research international 2020 (2020).

\bibitem{Zhang_2023}
Y.~Zhang, H.~Wang, D.~Butler, M.-S. To, J.~Avery, M.~L. Hull, G.~Carneiro, \href{http://dx.doi.org/10.1109/ISBI53787.2023.10230667}{Distilling missing modality knowledge from ultrasound for endometriosis diagnosis with magnetic resonance images}, in: 2023 IEEE 20th International Symposium on Biomedical Imaging (ISBI), IEEE, 2023.
\newblock \href {https://doi.org/10.1109/isbi53787.2023.10230667} {\path{doi:10.1109/isbi53787.2023.10230667}}.
\newline\urlprefix\url{http://dx.doi.org/10.1109/ISBI53787.2023.10230667}

\bibitem{goh2022CrowdLab}
H.~W. Goh, U.~Tkachenko, J.~Mueller, C.~C. Cleanlab, Crowdlab: Supervised learning to infer consensus labels and quality scores for data with multiple annotators, arXiv preprint arXiv:2210.06812 (2022).

\bibitem{complement_wilder}
B.~Wilder, E.~Horvitz, E.~Kamar, Learning to complement humans, in: Proceedings of the Twenty-Ninth International Joint Conference on Artificial Intelligence, IJCAI'20, 2021.

\bibitem{lu2021human}
Z.~Lu, M.~Yin, Human reliance on machine learning models when performance feedback is limited: Heuristics and risks, in: Proceedings of the 2021 CHI Conference on Human Factors in Computing Systems, 2021, pp. 1--16.

\bibitem{weitz2019you}
K.~Weitz, D.~Schiller, R.~Schlagowski, T.~Huber, E.~Andr{\'e}, " do you trust me?" increasing user-trust by integrating virtual agents in explainable ai interaction design, in: Proceedings of the 19th ACM International Conference on Intelligent Virtual Agents, 2019, pp. 7--9.

\bibitem{rosenfeld2018totally}
A.~Rosenfeld, M.~D. Solbach, J.~K. Tsotsos, Totally looks like-how humans compare, compared to machines, 2018, pp. 1961--1964.

\bibitem{serre2019deep}
T.~Serre, Deep learning: the good, the bad, and the ugly, Annual review of vision science 5 (2019) 399--426.

\bibitem{kamar2012combining}
E.~Kamar, S.~Hacker, E.~Horvitz, Combining human and machine intelligence in large-scale crowdsourcing., in: AAMAS, Vol.~12, 2012, pp. 467--474.

\bibitem{bansal2021most}
G.~Bansal, B.~Nushi, E.~Kamar, E.~Horvitz, D.~S. Weld, Is the most accurate ai the best teammate? optimizing ai for teamwork, in: AAAI, Vol.~35, 2021, pp. 11405--11414.

\bibitem{vodrahalli2022humans}
K.~Vodrahalli, R.~Daneshjou, T.~Gerstenberg, J.~Zou, Do humans trust advice more if it comes from ai? an analysis of human-ai interactions, in: Proceedings of the 2022 AAAI/ACM Conference on AI, Ethics, and Society, 2022, pp. 763--777.

\bibitem{humanIntheloop}
X.~Wu, L.~Xiao, Y.~Sun, J.~Zhang, T.~Ma, L.~He, \href{https://doi.org/10.1016/j.future.2022.05.014}{A survey of human-in-the-loop for machine learning}, Future Gener. Comput. Syst. 135~(C) (2022) 364–381.
\newblock \href {https://doi.org/10.1016/j.future.2022.05.014} {\path{doi:10.1016/j.future.2022.05.014}}.
\newline\urlprefix\url{https://doi.org/10.1016/j.future.2022.05.014}

\bibitem{pradier2021preferential}
M.~F. Pradier, J.~Zazo, S.~Parbhoo, R.~H. Perlis, M.~Zazzi, F.~Doshi-Velez, Preferential mixture-of-experts: Interpretable models that rely on human expertise as much as possible, AMIA Summits on Translational Science Proceedings 2021 (2021) 525.

\bibitem{cortes2016learning}
C.~Cortes, G.~DeSalvo, M.~Mohri, Learning with rejection, in: Algorithmic Learning Theory: 27th International Conference, ALT 2016, Bari, Italy, October 19-21, 2016, Proceedings 27, Springer, 2016, pp. 67--82.

\bibitem{madras2018predict}
D.~Madras, T.~Pitassi, R.~Zemel, Predict responsibly: improving fairness and accuracy by learning to defer, NeurIPS 31 (2018).

\bibitem{narasimhan2022post}
H.~Narasimhan, W.~Jitkrittum, A.~K. Menon, A.~Rawat, S.~Kumar, Post-hoc estimators for learning to defer to an expert, NeurIPS 35 (2022) 29292--29304.

\bibitem{raghu2019algorithmic}
M.~Raghu, K.~Blumer, G.~Corrado, J.~Kleinberg, Z.~Obermeyer, S.~Mullainathan, The algorithmic automation problem: Prediction, triage, and human effort, arXiv preprint arXiv:1903.12220 (2019).

\bibitem{okati2021differentiable}
N.~Okati, A.~De, M.~Rodriguez, Differentiable learning under triage, NeurIPS 34 (2021) 9140--9151.

\bibitem{verma2022calibration}
R.~Verma, D.~Barrej{\'o}n, E.~Nalisnick, On the calibration of learning to defer to multiple experts, in: Workshop on Human-Machine Collaboration and Teaming in International Confere of Machine Learning, 2022.

\bibitem{mao2023two}
A.~Mao, C.~Mohri, M.~Mohri, Y.~Zhong, Two-stage learning to defer with multiple experts, in: NeurIPS, 2023.

\bibitem{ijcai2022-344}
P.~Hemmer, S.~Schellhammer, M.~Vössing, J.~Jakubik, G.~Satzger, \href{https://doi.org/10.24963/ijcai.2022/344}{Forming effective human-ai teams: Building machine learning models that complement the capabilities of multiple experts}, International Joint Conferences on Artificial Intelligence Organization, 2022, pp. 2478--2484, main Track.
\newblock \href {https://doi.org/10.24963/ijcai.2022/344} {\path{doi:10.24963/ijcai.2022/344}}.
\newline\urlprefix\url{https://doi.org/10.24963/ijcai.2022/344}

\bibitem{steyvers2022bayesian}
M.~Steyvers, H.~Tejeda, G.~Kerrigan, P.~Smyth, Bayesian modeling of human--ai complementarity, Proceedings of the National Academy of Sciences 119~(11) (2022) e2111547119.

\bibitem{kerrigan2021combining}
G.~Kerrigan, P.~Smyth, M.~Steyvers, Combining human predictions with model probabilities via confusion matrices and calibration, NeurIPS 34 (2021) 4421--4434.

\bibitem{zhang2023learning}
Z.~Zhang, K.~Wells, G.~Carneiro, Learning to complement with multiple humans (lecomh): Integrating multi-rater and noisy-label learning into human-ai collaboration, arXiv preprint arXiv:2311.13172 (2023).

\bibitem{liu2023humans}
M.~Liu, J.~Wei, Y.~Liu, J.~Davis, Do humans and machines have the same eyes? human-machine perceptual differences on image classification, arXiv preprint arXiv:2304.08733 (2023).

\bibitem{Dou2020unpaired}
Q.~Dou, Q.~Liu, P.~A. Heng, B.~Glocker, Unpaired multi-modal segmentation via knowledge distillation, in: IEEE Transactions on Medical Imaging, 2020.

\bibitem{Monteiro2020stochastic}
M.~Monteiro, L.~Le~Folgoc, D.~Coelho~de Castro, N.~Pawlowski, B.~Marques, K.~Kamnitsas, M.~van~der Wilk, B.~Glocker, Stochastic segmentation networks: Modelling spatially correlated aleatoric uncertainty, Advances in Neural Information Processing Systems 33 (2020) 12756--12767.

\bibitem{Han2021trusted}
Z.~Han, C.~Zhang, H.~Fu, J.~T. Zhou, Trusted multi-view classification, arXiv preprint arXiv:2102.02051 (2021).

\bibitem{Wang2022uncertainty}
H.~Wang, J.~Zhang, Y.~Chen, C.~Ma, J.~Avery, L.~Hull, G.~Carneiro, Uncertainty-aware multi-modal learning via cross-modal random network prediction, in: European Conference on Computer Vision, Springer, 2022, pp. 200--217.

\bibitem{wang2024enhancing}
H.~Wang, C.~Ma, Y.~Liu, Y.~Chen, Y.~Tian, J.~Avery, L.~Hull, G.~Carneiro, Enhancing multi-modal learning: Meta-learned cross-modal knowledge distillation for handling missing modalities, arXiv preprint arXiv:2405.07155 (2024).

\bibitem{wang2023learnable}
H.~Wang, C.~Ma, J.~Zhang, Y.~Zhang, J.~Avery, L.~Hull, G.~Carneiro, Learnable cross-modal knowledge distillation for multi-modal learning with missing modality, in: International Conference on Medical Image Computing and Computer-Assisted Intervention, Springer, 2023, pp. 216--226.

\bibitem{wang2023multi}
H.~Wang, Y.~Chen, C.~Ma, J.~Avery, L.~Hull, G.~Carneiro, Multi-modal learning with missing modality via shared-specific feature modelling, in: Proceedings of the IEEE/CVF Conference on Computer Vision and Pattern Recognition, 2023, pp. 15878--15887.

\bibitem{Wang2020deep}
Y.~Wang, W.~Huang, F.~Sun, T.~Xu, Y.~Rong, J.~Huang, Deep multimodal fusion by channel exchanging, Advances in Neural Information Processing Systems 33 (2020) 4835--4845.

\bibitem{Patrick2020multi}
M.~Patrick, Y.~M. Asano, P.~Kuznetsova, R.~Fong, J.~F. Henriques, G.~Zweig, A.~Vedaldi, Multi-modal self-supervision from generalized data transformations, arXiv preprint arXiv:2003.04298 (2020).

\bibitem{Patrick2021space}
M.~Patrick, P.-Y. Huang, I.~Misra, F.~Metze, A.~Vedaldi, Y.~M. Asano, J.~F. Henriques, Space-time crop \& attend: Improving cross-modal video representation learning, in: Proceedings of the IEEE/CVF International Conference on Computer Vision, 2021, pp. 10560--10572.

\bibitem{Chen2021localizing}
H.~Chen, W.~Xie, T.~Afouras, A.~Nagrani, A.~Vedaldi, A.~Zisserman, Localizing visual sounds the hard way, in: Proceedings of the IEEE/CVF Conference on Computer Vision and Pattern Recognition, 2021, pp. 16867--16876.

\bibitem{Jia2020semi}
X.~Jia, X.-Y. Jing, X.~Zhu, S.~Chen, B.~Du, Z.~Cai, Z.~He, D.~Yue, Semi-supervised multi-view deep discriminant representation learning, IEEE transactions on pattern analysis and machine intelligence 43~(7) (2020) 2496--2509.

\bibitem{Lee2018diverse}
H.-Y. Lee, H.-Y. Tseng, J.-B. Huang, M.~Singh, M.-H. Yang, Diverse image-to-image translation via disentangled representations, in: Proceedings of the European conference on computer vision (ECCV), 2018, pp. 35--51.

\bibitem{Liu2022learning}
X.~Liu, P.~Sanchez, S.~Thermos, A.~Q. O’Neil, S.~A. Tsaftaris, Learning disentangled representations in the imaging domain, Medical Image Analysis (2022) 102516.

\bibitem{zhou2012ensemble}
Z.-H. Zhou, Ensemble methods: foundations and algorithms, CRC press, 2012.

\bibitem{rodrigues2014gaussian}
F.~Rodrigues, F.~Pereira, B.~Ribeiro, Gaussian process classification and active learning with multiple annotators, in: ICML, PMLR, 2014, pp. 433--441.

\bibitem{raykar2010learning}
V.~C. Raykar, S.~Yu, L.~H. Zhao, G.~H. Valadez, C.~Florin, L.~Bogoni, L.~Moy, Learning from crowds., Journal of machine learning research 11~(4) (2010).

\bibitem{rodrigues2018deep}
F.~Rodrigues, F.~Pereira, Deep learning from crowds, Vol.~32, 2018.

\bibitem{chen2021structured}
Z.~Chen, H.~Wang, H.~Sun, P.~Chen, T.~Han, X.~Liu, J.~Yang, Structured probabilistic end-to-end learning from crowds, 2021, pp. 1512--1518.

\bibitem{guan2018said}
M.~Guan, V.~Gulshan, A.~Dai, G.~Hinton, Who said what: Modeling individual labelers improves classification, in: AAAI, Vol.~32, 2018.

\bibitem{crowdlab}
H.~W. Goh, U.~Tkachenko, J.~Mueller, Crowdlab: Supervised learning to infer consensus labels and quality scores for data with multiple annotators (2023).
\newblock \href {http://arxiv.org/abs/2210.06812} {\path{arXiv:2210.06812}}.

\bibitem{he2022masked}
K.~He, X.~Chen, S.~Xie, Y.~Li, P.~Doll{\'a}r, R.~Girshick, Masked autoencoders are scalable vision learners, in: Proceedings of the IEEE/CVF conference on computer vision and pattern recognition, 2022, pp. 16000--16009.

\bibitem{dosovitskiy2020image}
A.~Dosovitskiy, et~al., An image is worth 16x16 words: Transformers for image recognition at scale, arXiv preprint arXiv:2010.11929 (2020).

\bibitem{hara2017learning}
K.~Hara, H.~Kataoka, Y.~Satoh, Learning spatio-temporal features with 3d residual networks for action recognition, in: Proceedings of the IEEE international conference on computer vision workshops, 2017, pp. 3154--3160.

\bibitem{loshchilov2016sgdr}
I.~Loshchilov, F.~Hutter, Sgdr: Stochastic gradient descent with warm restarts, arXiv preprint arXiv:1608.03983 (2016).

\bibitem{Feng2022SSR}
C.~Feng, G.~Tzimiropoulos, I.~Patras, \href{https://bmvc2022.mpi-inf.mpg.de/372/}{{SSR:} an efficient and robust framework for learning with unknown label noise}, in: 33rd British Machine Vision Conference 2022, {BMVC} 2022, London, UK, November 21-24, 2022, {BMVA} Press, 2022, p. 372.
\newline\urlprefix\url{https://bmvc2022.mpi-inf.mpg.de/372/}

\bibitem{xiao2023promix}
R.~Xiao, Y.~Dong, H.~Wang, L.~Feng, R.~Wu, G.~Chen, J.~Zhao, Promix: combating label noise via maximizing clean sample utility, in: Proceedings of the Thirty-Second International Joint Conference on Artificial Intelligence, IJCAI-23, E. Elkind, Ed. International Joint Conferences on Artificial Intelligence Organization, Vol.~8, 2023, pp. 4442--4450.

\end{thebibliography}

\end{document}